\documentclass[11pt]{article}
\usepackage{abstract,amsmath,amssymb,latexsym}
\usepackage{enumitem,epsf}
\usepackage{fullpage,tikz,float}
\usepackage[numbers]{natbib}
\usepackage[pdftex,colorlinks]{hyperref}
\usepackage{macros}

\newif\ifdraft
\drafttrue

\title{Depth Separation for Neural Networks}

\author{
	\vspace{1cm}
  Amit Daniely\thanks{Google Brain} 
}

\begin{document}
\maketitle
\maketitle

\begin{abstract}
Let $f:\sphere^{d-1}\times \sphere^{d-1}\to\reals$ be a function of the form $f(\x,\x') = g(\inner{\x,\x'})$ for $g:[-1,1]\to \reals$. We give a simple proof that shows that poly-size depth two neural networks with (exponentially) bounded weights cannot approximate $f$ whenever $g$ cannot be approximated by a low degree polynomial. Moreover, for many $g$'s, such as $g(x)=\sin(\pi d^3x)$, the number of neurons must be $2^{\Omega\left(d\log(d)\right)}$. Furthermore, the result holds w.r.t.\ the uniform distribution on $\sphere^{d-1}\times \sphere^{d-1}$.
As many functions of the above form can be well approximated by poly-size depth three networks with poly-bounded weights, this establishes a separation between depth two and depth three networks w.r.t.\ the uniform distribution on $\sphere^{d-1}\times \sphere^{d-1}$.
\end{abstract}

\section{Introduction and main result}
Many aspects of the expressive power of neural networks has been studied over the years. In particular, separation for deep networks~\citep{telgarsky2015representation, safran2016depth}, expressive power of depth two networks~\citep{Cybenko88, hornik1989multilayer, funahashi1989approximate, barron1994approximation}, and more~\citep{delalleau2011shallow, cohen2016expressive}.
We focus on the basic setting of depth $2$ versus depth $3$ networks. We ask what functions are expressible (or well approximated) by poly-sized depth-$3$ networks, but cannot be approximated by an exponential size depth-$2$ network. 

Two recent papers \citep{martens2013representational, eldan2016power} addressed this issue. Both papers presented a specific function $f:\reals^d\to\reals$ and a distribution $\cd$ on $\reals^d$ such that $f$ can be approximated w.r.t.\ $\cd$ by a $\poly(d)$-size depth $3$ network, but not by a $\poly(d)$-size depth $2$ network. In \citet{martens2013representational} this was shown for $f$ being the inner product mod 2 and $\cd$ being the uniform distribution on $\{0,1\}^d\times \{0,1\}^d$. In \citet{eldan2016power} it was shown for a different (radial) function and some (unbounded) distribution.

We extend the above results and prove a similar result for an explicit and rich family of functions, and w.r.t.\ the uniform distribution on $\sphere^{d-1}\times\sphere^{d-1}$. In addition, our lower bound on the number of required neurons is stronger: while previous papers showed that the number of neurons has to be exponential in $d$, we show exponential dependency on $d\log(d)$. Last, our proof is short, direct and is based only on basic Harmonic analysis over the sphere. In contrast, \citet{eldan2016power}'s proof is rather lengthy and requires advanced technical tools such as tempered distributions, while \citet{martens2013representational} relied on the discrepancy of the inner product function mod 2. On the other hand, \citet{eldan2016power} do not put any restriction on the magnitude of the weights, while we and \citet{martens2013representational} do require a mild (exponential) bound.

Let us fix an activation function $\sigma:\reals\to\reals$. For $\x\in \reals^n$ we denote $\sigma(\x) = (\sigma(x_1),\ldots,\sigma(x_n))$.
We say that $F:\sphere^{d-1}\times\sphere^{d-1}\to\reals$ can be implemented by a depth-$2$ $\sigma$-network of width $r$ and weights bounded by $B$ if
\[
F(\x,\x') = w^T_2\sigma(W_1\x + W'_1\x' + b_1)+b_2~,
\]
where $W_1,W'_1\in [-B,B]^{r\times d}, w_2\in [-B,B]^r$, $b_1\in [-B,B]^r$ and $b_2\in [-B,B]$.
Similarly, $F:\sphere^{d-1}\times\sphere^{d-1}\to\reals$ can be implemented by a depth-$3$ $\sigma$-network of width $r$ and weights bounded by $B$ if
\[
F(\x,\x') = w^T_3\sigma(W_2\sigma(W_1\x + W'_1\x' + b_1)+b_2) + b_3
\]
for $W_1,W'_1\in [-B,B]^{r\times d}, W_2\in [-B,B]^{r\times r}, w_3\in [-B,B]^r$, $b_1, b_2\in [-B,B]^r$ and $b_3\in [-B,B]$.
Denote
\[
N_{d,n}=\binom{d+n-1}{d-1}-\binom{d+n-3}{d-1}=\frac{(2n+d-2)(n+d-3)!}{n!(d-2)!}~.
\]
Let $\mu_d$ be the probability measure on $[-1,1]$ given by $d\mu_d(x) = \frac{\Gamma\left(\frac{d}{2}\right)}{\sqrt{\pi}\Gamma\left(\frac{d-1}{2}\right)}(1-x^2)^{\frac{d-3}{2}}dx$ and define
\[
A_{n,d}(f) = \min_{p\text{ is degree }n-1\text{ polynomial}}\|f-p\|_{L^2(\mu_d)}
\]
Our main theorem shows that if $A_{n,d}(f)$ is large then $(\x,\x')\mapsto f(\inner{\x,\x'})$ cannot be approximated by a small depth-$2$ network.

\begin{theorem}[main]\label{thm:main}
Let $N:\sphere^{d-1}\times\sphere^{d-1}\to\reals$ be any function implemented by a depth-$2$ $\sigma$-network of width $r$, with weights bounded by $B$. Let $f:[-1,1]\to\reals$ and define $F:\sphere^{d-1}\times\sphere^{d-1}\to\reals$ by $F(\x,\x') = f(\inner{\x,\x'})$.
Then, for all $n$,
\[
\|N - F\|_{L^2(\sphere^{d-1}\times\sphere^{d-1})} \ge  A_{n,d}(f)\left(A_{n,d}(f) - \frac{2rB\max_{|x|\le \sqrt{4d}B+B}|\sigma(x)|+2B}{\sqrt{N_{d,n}}}\right)
\]
\end{theorem}

\begin{example}\label{exam:sine}
Let us consider the case that $\sigma(x)=\max(0,x)$ is the ReLU function, $f(x) = \sin(\pi d^{3}x)$, $n=d^2$ and $B=2^d$. In this case, lemma \ref{lem:sine} implies that $A_{n,d}(f) \ge \frac{1}{5e\pi}$. Hence, to have $\frac{1}{50e^2\pi^2}$-approximation of $F$, the number of hidden neuorons has to be at least,
\[
\frac{\sqrt{N_{d,d^2}}}{20e\pi 2^{2d}(1+\sqrt{4d}) + 2^{d+1}} = 2^{\Omega\left(d\log(d)\right)}
\]
On the other hand, corollary \ref{cor:3_layer_apx} implies that $F$ can be $\epsilon$-approximated by a ReLU network of depth $3$, width $\frac{16\pi d^5}{\epsilon}$ and weights bounded by $2\pi d^3$
\end{example}

\section{Proofs}
Throughout, we fix a dimension $d$. All functions $f:\sphere^{d-1}\to\reals$ and $f:\sphere^{d-1}\times\sphere^{d-1}\to\reals$ will be assumed to be square integrable w.r.t.\ the uniform measure. Likewise, functions $f:[-1,1]\to\reals$ and $f:[-1,1]\times [-1,1]\to\reals$ will be assumed to be square integrable w.r.t.\ $\mu_d$ or $\mu_d\times\mu_d$. Norms and inner products of such functions are of the corresponding $L^2$ spaces. We will use the fact that $\mu_d$ is the probability measure on $[-1,1]$ that is obtained by pushing forward the uniform measure on $\sphere^{d-1}$ via the function $\x\mapsto x_1$. We denote by $\cp_{n}:L^2(\mu_d)\to L^2(\mu_d)$ the projection on the complement of the space of degree $\le n-1$ polynomials. Note that $A_{n,d}(f)=\|\cp_{n,d}f\|_{L^2(\mu_d)}$.

\subsection{Some Harmonic Analysis on the Sphere}
The {\em $d$ dimensional Legendre polynomials} are the sequence of polynomials over $[-1,1]$ defined by the recursion formula
\begin{eqnarray*}
& P_{n}(x)=\frac{2n+d-4}{n+d-3}xP_{n-1}(x) - \frac{n-1}{n+d-3}P_{n-2}(x)\\
& P_{0}\equiv 1,\;P_{1}(x)=x
\end{eqnarray*}
We also define $h_{n}:S^{d-1}\times S^{d-1}\to\reals$ by $h_{n}(\x,\x') = \sqrt{N_{d,n}}P_{n}(\inner{\x,\x'})$, and for $\x\in S^{d-1}$ we denote $L^{\x}_{n}(\x') = h_{n}(\x,\x')$. We will make use of the following properties of the Legendre polynomials.
\begin{proposition}[e.g.\ \cite{AtkinsonHa12} chapters 1 and 2]\label{prop:Legendre}\
\begin{enumerate}
\item For every $d\ge 2$, the sequence $\{\sqrt{N_{d,n}}P_{n}\}$ is orthonormal basis of the Hilbert space $L^2\left(\mu_d\right)$.
\item For every $n$, $||P_{n}||_\infty=1$ and $P_{n}(1)=1$.
\item $\inner{L_{i}^\x,L_{j}^{\x'}} = P_{i}(\inner{\x,\x'})\delta_{ij}$. 
\end{enumerate}
\end{proposition}

\subsection{Main Result}
We say that $f:\sphere^{d-1}\times\sphere^{d-1}\to\reals$ is an {\em inner product function} if it has the form $f(\x,\x') = \phi(\inner{\x,\x'})$ for some function $\phi:[-1,1]\to\reals$.
Let $\ch_d\subset L^2(\sphere^{d-1}\times\sphere^{d-1})$ be the space of inner product functions. We note that
\[
\|f\|^2 = \E_{\x}\E_{\x'}\phi^2(\inner{\x,\x'})  =  \E_\x \|\phi\|^2 = \|\phi\|^2
\]
Hence, the correspondence $\phi\leftrightarrow f$ defines an isomorphism of Hilbert spaces between $L^2(\mu_d)$ and $\ch_d$. In particular, the orthonormal basis $\{\sqrt{N_{d,n}}P_n\}_{n=0}^\infty$ is mapped to $\{h_{n}\}_{n=0}^\infty$. In particular,
\[
\cp_n\left(\sum_{i=0}^\infty \alpha_ih_i \right) = \sum_{i=n}^\infty \alpha_ih_i 
\]
Let $\bv,\bv'\in \sphere^{d-1}$. We say that $f:\sphere^{d-1}\times\sphere^{d-1}\to\reals$ is {\em $(\bv,\bv')$-separable} if it has the form $f(\x,\x') = \psi(\inner{\bv,\x},\inner{\bv',\x'})$ for some $\psi:[-1,1]^2\to\reals$. We note that each neuron implements a separable function. 
Let $\ch_{\bv,\bv'}\subset L^2(\sphere^{d-1}\times\sphere^{d-1})$ be the space of $(\bv,\bv')$-separable functions.
We note that
\[
\|f\|^2 = \E_{\x,\x'}\psi^2(\inner{\bv,\x}, \inner{\bv',\x'})  =  \|\psi\|^2
\]
Hence, the correspondence $\psi\leftrightarrow f$ defines an isomorphism of Hilbert spaces between $L^2(\mu_d\times \mu_d)$ and $\ch_{\bv,\bv'}$. In particular, the orthonormal basis $\{\sqrt{N_{d,n}}P_n \otimes \sqrt{N_{d,m}}P_m\}_{n,m=0}^\infty$ is mapped to $\{L_n^{\bv}\otimes L_n^{\bv'}\}_{n,m=0}^\infty$.

The following theorem implies theorem \ref{thm:main}, as under the conditions of theorem \ref{thm:main}, any hidden neuron implement a separable function with norm at most $B\max_{|x|\le \sqrt{4d}B+B}|\sigma(x)|$, and the bias term is a separable function with norm at most $B$.

\begin{theorem}
Let $f:\sphere^{d-1}\times\sphere^{d-1}\to \reals$ be an inner product function and let $g_1,\ldots,g_r:\sphere^{d-1}\times\sphere^{d-1}\to \reals$ be separable functions. Then
\begin{equation}\label{eq:appr_lower}
\left\|f-\sum_{i=1}^rg_i\right\|^2 \ge \|\cp_nf\|\left(\|\cp_nf\| - \frac{2\sum_{i=1}^r\|g_i\|}{\sqrt{N_{d,n}}}\right)
\end{equation}
\end{theorem}
\proof
We note that
\begin{eqnarray}\label{eq:basic}
\E_{\x,\x'}h_n(\x,\x')L_{i}^{\bv}(\x)L_{j}^{\bv'}(\x') &=& \E_{\x}L_{i}^{\bv}(\x)\E_{\x'}h_n(\x,\x')L_{j}^{\bv'}(\x') \nonumber
\\
&=& \E_{\x}L_{i}^{\bv}(\x)\E_{\x'}L^{\x}_n(\x')L_{j}^{\bv'}(\x') \nonumber
\\
&=& \delta_{nj}\E_{\x}L_{i}^{\bv}(\x)P_n(\inner{\x,\bv'})
\\
&=& \frac{\delta_{nj}}{\sqrt{N_{d,n}}}\E_{\x}L_{i}^{\bv}(\x)L_{n}^{\bv'}(\x) \nonumber
\\
&=& \frac{\delta_{nj}\delta_{ni} P_n(\inner{\bv,\bv'}) }{\sqrt{N_{d,n}}} \nonumber
\end{eqnarray}
Suppose now that $f = \sum_{i=n}^{\infty} \alpha_ih_i$ and suppose that $g = \sum^r_{j=1} g_j$ where each $g_j$ depends only on $\inner{\bv_j,\x},\inner{\bv'_j,\x'}$ for some $\bv_j,\bv'_j\in S^{d-1}$. 
Write $g_j(\x,\x')=\sum_{k,l=0}^\infty \beta^j_{k,l}L^{\bv_j}_k(\x)L^{\bv'_j}_k(\x')$. By equation \eqref{eq:basic}, $L^{\bv_j}_k(\x)L^{\bv'_j}_l(\x')$ is orthogonal to $f$ whever $k\ne l$. Hence, if we replace each $g_j$ with $\sum_{k=0}^\infty \beta^j_{k,k}L^{\bv_j}_k(\x)L^{\bv'_j}_k(\x')$, the l.h.s.\ of \eqref{eq:appr_lower} does not increase. Likewise, the r.h.s.\ does not decrease.
Hence, we can assume w.l.o.g.\ that each $g_j$ is of the form $g_j(\x,\x')=\sum_{i=0}^\infty \beta^j_{i}L^{\bv_j}_i(\x)L^{\bv'_j}_i(\x')$. Now, using \eqref{eq:basic} again, we have that
\begin{eqnarray*}
\|f-g\|^2  &=& \sum_{i=0}^\infty \left\|\alpha_i h_i - \sum_{j=1}^r \beta^j_i  L^{\bv_j}_i \otimes L^{\bv'_j}_i\right\|^2
\\
&\ge & \sum_{i=n}^\infty \left\|\alpha_i h_i - \sum_{j=1}^r \beta^j_i  L^{\bv_j}_i \otimes L^{\bv'_j}_i\right\|^2
\\
&\ge & \sum_{i=n}^\infty \alpha_i^2  - 2\sum_{i=n}^\infty\sum_{j=1}^r\inner{\alpha_i h_i, \beta^j_i  L^{\bv_j}_i \otimes L^{\bv'_j}_i}
\\
&=& \|\cp_nf\|^2-2\sum_{i=n}^\infty\sum_{j=1}^r \frac{\beta^j_{i}\alpha_i  P_i(\inner{\bv_j,\bv'_j})}{\sqrt{N_{d,k}}}
\\
&\ge& \|\cp_nf\|^2-2\sum^r_{j=1}\sum_{i=n}^\infty \frac{|\beta^j_{i}| |\alpha_i|}{\sqrt{N_{d,n}}}
\\
&\ge& \|\cp_nf\|^2-2\sum^r_{j=1}\frac{1}{\sqrt{N_{d,n}}}\sqrt{\sum_{i=n}^\infty |\beta^j_i|^2}\sqrt{\sum_{i=n}^\infty |\alpha_i|^2}
\\
&\ge & \|\cp_nf\|^2 - \frac{2\|\cp_nf\|\sum_{j=1}^r \|g_j\|}{\sqrt{N_{d,n}}}
\end{eqnarray*}
\proofbox

\subsection{Approximating the cosine function}
\begin{lemma}\label{lem:sine}
Define $g_{d,m}(x) = \sin\left(\pi \sqrt{d}m x\right)$. Then, for any $d\ge d_0$, for a universal constant $d_0>0$, and for any degree $k$ polynomial $p$ we have
\[
\int_{-1}^1 (g_{d,m}(x)-p(x))^2d\mu_d(x) \ge \frac{m-k}{4e\pi m} 
\]
\end{lemma}
\proof We have that (e.g.\ \cite{AtkinsonHa12}) $d\mu_d(x) = \frac{\Gamma\left(\frac{d}{2}\right)}{\sqrt{\pi}\Gamma\left(\frac{d-1}{2}\right)}(1-x^2)^{\frac{d-3}{2}}dx$. Likewise, for large enough $d$ and $|x|< \frac{1}{\sqrt{d}}$ we have $1-x^2 \ge e^{-2x^2}\ge e^{-\frac{2}{d}}$ and hence $(1-x^2)^{\frac{d-3}{2}} \ge e^{-\frac{d-3}{d}}\ge e^{-1}$.
Likewise, since $\frac{\Gamma\left(\frac{d}{2}\right)}{\Gamma\left(\frac{d-1}{2}\right)}\sim \sqrt{\frac{d}{2}}$, we have that for large enough $d$ and $|x|\le\frac{1}{\sqrt{d}}$, $d\mu_d(x)\ge \frac{\sqrt{d}}{2e\pi}$. Hence, for $f\ge 0$ we have
\[
\int_{-1}^1 f(x)d\mu_d(x) \ge \int_{-d^{-\frac{1}{2}}}^{d^{-\frac{1}{2}}} f(x)d\mu_d(x) \ge  \frac{\sqrt{d}}{2e\pi}\int_{-d^{-\frac{1}{2}}}^{d^{-\frac{1}{2}}}f(x)dx = \frac{1}{2e\pi}\int_{-1}^{1}f\left(\frac{t}{\sqrt{d}}\right)dt
\]
Applying this equation for $f = g_{d,m} - p$ we get that
\[
\int_{-1}^1 (g_{d,m}(x)-p(x))^2d\mu_d(x) \ge  \frac{1}{2e\pi}\int_{-1}^1 \left(\sin(\pi m x) - q(x)\right)^2dx
\]
Where $q(x):=p\left(\frac{x}{\sqrt{d}}\right)$.
Now, in the $2m$ segments $I_i = \left(-1+\frac{i-1}{m},-1 + \frac{i}{m}\right),\;\;i\in [2m]$ we have at least $m - k$ segments on which $x\mapsto \sin(\pi m x)$ and $q$ do not change signs and have opposite signs. On each of these intervals we have $\int_I \left(\sin(\pi m x) - q(x)\right)^2dx \ge \int^{\frac{1}{m}}_{0} \sin^2(\pi m x)dx =\frac{1}{2m}$.
\proofbox

\begin{lemma}[e.g.\ \cite{eldan2016power}]\label{lem:one_dim_apx}
Let $\sigma(x)=\max(x,0)$ be the ReLU activation, $f:[-R,R]\to\reals$ an $L$-Lipschitz function, and $\epsilon>0$. There is a function
\[
g(x) = f(0) + \sum_{i=1}^m \alpha_i\sigma(\gamma_i x-\beta_i)
\]
for which $\|g-f\|_\infty\le \epsilon$. Furthermore, $m\le \frac{2RL}{\epsilon}$, $|\beta_i|\le R$, $|\alpha_i|\le 2L$, $\gamma_i\in \{-1,1\}$, and $g$ is $L$-Lipschitz on all $\reals$. 
\end{lemma}

\begin{corollary}\label{cor:3_layer_apx}
Let $f:[-1,1]\to [-1,1]$ be an $L$-Lipschitz function and let $\epsilon>0$. Define $F:\sphere^{d-1}\times\sphere^{d-1}\to [-1,1]$ by $F(\x,\x') = f(\inner{\x,\x'})$. There is a function $G:\sphere^{d-1}\times\sphere^{d-1}\to [-1,1]$ that satisfies $\|F-G\|_\infty\le\epsilon$ and furthermore $G$ can be implemented by a depth-$3$ ReLU network of width $\frac{16d^2L}{\epsilon}$ and weights bounded by $\max(4,2L)$
\end{corollary}
\proof
By Lemma \ref{lem:one_dim_apx} there is a depth-$2$ network $\cn_\text{square}$ that calculates $\frac{x^2}{2}$ in $[-2,2]$, with an error of $\frac{\epsilon}{2dL}$ and has width at most $\frac{16dL}{\epsilon}$ and hidden layer weights bounded by $2$, and prediction layer weights bounded by $4$. For each $i\in [d]$ we can compose the linear function $(\x,\x')\mapsto x_i+x'_i$ with $\cn_\text{square}$ to get a depth-2 network $\cn_{i}$ that calculates $\frac{(x_i+x'_i)^2}{2}$ with an error of $\frac{\epsilon}{2dL}$ and has the same width and weight bound as $\cn_\text{square}$. Summing the networks $\cn_i$ and subtracting $1$ results with a depth-2 network $\cn_{\text{inner}}$ that calculates $\inner{\x,\x'}$ with an error of $\frac{\epsilon}{2L}$ and has width $\frac{16d^2L}{\epsilon}$ and hidden layer weights bounded by $2$, and prediction layer weights bounded by $4$.

Now, again by lemma \ref{lem:one_dim_apx} there is a depth-$2$ network $\cn_f$ that calculates $f$ in $[-1,1]$, with an error of $\frac{\epsilon}{2}$, has width at most $\frac{2L}{\epsilon}$, hidden layer weights bounded by $1$ and prediction layer weights bounded by $2L$, and is $L$-Lipschitz. Finally, consider the depth-$3$ network $\cn_F$ that is the composition of $\cn_{\text{inner}}$ and $\cn_f$. $\cn_F$ has width at most $\frac{16d^2L}{\epsilon}$ weight bound of $\max(4,2L)$, and it satisfies
\begin{eqnarray*}
|\cn_F(\x,\x') - F(\x,\x')| &=& |\cn_f(\cn_{\text{inner}}(\x,\x')) - f(\inner{\x,\x'})|
\\
&\le& |\cn_f(\cn_{\text{inner}}(\x,\x')) - \cn_f(\inner{\x,\x'})| + |\cn_f(\inner{\x,\x'})- f(\inner{\x,\x'})|
\\
&\le& L|\cn_{\text{inner}}(\x,\x') -\inner{\x,\x'}| + \frac{\epsilon}{2}
\\
&\le& L\frac{\epsilon}{2L} + \frac{\epsilon}{2} = \epsilon
\end{eqnarray*}

\proofbox

\bibliography{bib}
\end{document}